\documentclass[11pt,a4paper]{article}
\usepackage{authblk}
\usepackage[hyphens,spaces,obeyspaces]{url}

\usepackage[hyperref]{emnlp2018}
\usepackage{times}
\usepackage{latexsym}
\usepackage{url}
\usepackage{booktabs}
\usepackage{amsmath}
\usepackage{multirow}
\usepackage{url}
\usepackage{float}
\usepackage{graphicx}
\usepackage{tikz}
\usepackage[normalem]{ulem}

\aclfinalcopy % Uncomment this line for the final submission

\title{Towards Exploiting Background Knowledge for Building Conversation Systems}

\author[1,2]{\textbf{Nikita Moghe}}
\author[1]{\textbf{Siddhartha Arora}}
\author[1]{\textbf{Suman Banerjee}}
\author[1,2] {\textbf{Mitesh M. Khapra}}
\affil[1]{Department of Computer Science and Engineering, Indian Institute of Technology Madras}
\affil[2]{Robert Bosch Centre for Data Science and AI (RBC-DSAI), \protect\\ Indian Institute of Technology Madras}
\affil[ ]{\tt{\{nikitavam,sidarora,suman,miteshk\}}@cse.iitm.ac.in}

\date{}

\begin{document}
\maketitle
\begin{abstract}

Existing dialog datasets contain a sequence of utterances and responses without any explicit background knowledge associated with them. This has resulted in the development of models which treat conversation as a sequence-to-sequence generation task (\textit{i.e.}, given a sequence of utterances generate the response sequence). This is not only an overly simplistic view of conversation but it is also emphatically different from the way humans converse by heavily relying on their background knowledge about the topic (as opposed to simply relying on the previous sequence of utterances). For example, it is common for humans to (involuntarily) produce utterances which are copied or suitably modified from background articles they have read about the topic. To facilitate the development of such natural conversation models which mimic the human process of conversing, we create a new dataset containing movie chats wherein each response is explicitly generated by copying and/or modifying sentences from unstructured background knowledge such as plots, comments and reviews about the movie.  We establish baseline results on this dataset (90K utterances from 9K conversations) using three different models: (i) pure generation based models which ignore the background knowledge (ii) generation based models which learn to copy information from the background knowledge when required and (iii) span prediction based models which predict the appropriate response span in the background knowledge.  
\end{abstract}

\section{Introduction}

Background knowledge plays a very important role in human conversations. For example, to have a meaningful conversation about a movie, one uses their knowledge about the plot, reviews, comments and facts about the movie.
A typical conversation involves recalling important points from this background knowledge and producing them appropriately in the context of the  conversation. However, most existing large scale datasets \cite{ubuntu,ritter2010,HRED} simply contain a sequence of utterances and responses without any \textit{explicit} background knowledge associated with them. This has led to the development of models which treat conversation as a simple sequence-to-sequence generation task and often produce output which is both syntactically incorrect and incoherent (off topic). To make conversations more coherent, there is an increasing interest in integrating structured and unstructured knowledge sources with neural conversation models. While there are already some works in this direction \cite{trainable_task_oriented,dst_review,lowe2015,KGNCM} which try to integrate external knowledge sources with existing datasets, we believe that building new datasets where the utterances are \textit{explicitly} linked to external background knowledge will further facilitate the development of such background aware conversation models. 

\begin{figure*}
\label{dataset_example}
\centering
\resizebox{12.5cm}{7.7cm}{
\begin{tikzpicture}
\small
\node[draw,text width=3.5cm] at (-7.7,16) {...  The lab works on spiders and has even managed to create new species of spiders through genetic manipulation.\textcolor{blue}{\uline{ While Peter is taking photographs of Mary Jane for the school newspaper, one of these new spiders lands on his hand and bites him }} Peter comes home feeling ill and immediately goes to bed. ...};

\node[font=\bfseries] at (-7.7,18.7) {Plot};

\node[draw,text width=3.5 cm] at (-7.7,10.7)  {... \textcolor{blue} { \uline{I thoroughly enjoyed ``Spider-Man''}} which I saw in a screening. I thought the movie was very engrossing.  Director Sam Raimi kept the action quotient high, but also emphasized the human element of the story.  Tobey was brilliant as a gawky teenager...
};

\node[font=\bfseries] at (-7.7,13.2) {Review};

\node[draw,text width=7.1cm] at (-1.75,13.5) {Speaker 1(N): Which is your favourite character?\\
\vspace{2mm}
Speaker 2(C): My favorite character was Tobey Maguire. \\\vspace{2mm}

Speaker 1(N): I thought he did an excellent job as peter parker, I didn't see what it was that turned him into Spider-Man though.\\\vspace{2mm}

Speaker 2(P): Well this happens while Peter is taking photographs of Mary Jane for the school newspaper, one of these new spiders lands on his hand and bites him. \\\vspace{2mm}

Speaker 1 (N): I see. I was very excited to see this film and it did not disappoint! \\\vspace{2mm}

Speaker 2(R): I agree,  I thoroughly enjoyed ``Spider-Man''\\\vspace{2mm}

Speaker 1(N): I loved that they stayed true to the comic.\\\vspace{2mm}

Speaker 2(C): Yeah, it was a  really great comic book adaptation\\\vspace{2mm}

Speaker 1(N): The movie is a great life lesson on balancing power.\\\vspace{2mm}

Speaker 2(F): That is my most favorite line in the movie, ``With great power comes great responsibility.''\vspace{2mm}

};

\node[font=\bfseries] at (-1.75,18.7){Movie: Spider-Man};

\node[draw,text width=3.5cm] at (4.4,16) {... Crazy attention to detail. \textcolor{blue} {\uline{My favorite character was Tobey Maguire.}} I can't get over the ``I'm gonna kill you dead'' line. It was too heavily reliant on constant light-hearted humor. However the constant joking around kinda bogged it down for me. \textcolor{blue} { \uline{A really great comic book adaptation.}}  ....};
\node[font=\bfseries] at (4.25,18.6) {Comments};

\node[text width=0.002cm] at (2.4,10.58) {\begin{tabular}{|l|l|}
\hline
%Collection                                               & \$403,706,375                                                                                                                     \\ \hline
Awards                                                   & \begin{tabular}[c]{@{}l@{}}Golden Trailer\\ Awards 2002 \\ \end{tabular}                 \\ \hline
Taglines                                                 & \begin{tabular}[c]{@{}l@{}} \textcolor{blue} {\uline{With great}}\\ \textcolor{blue} {\uline{power comes}}\\ \textcolor{blue} {\uline{great}} \\ \textcolor{blue} {\uline{responsibility.}} \\ Get Ready For\\ Spidey !\end{tabular} \\ \hline
\begin{tabular}[c]{@{}l@{}}Similar\\ Movies\end{tabular} & {\color[HTML]{000000} \begin{tabular}[c]{@{}l@{}}Iron Man\\ Spider-Man 2 \end{tabular}}                                             \\ \hline
\end{tabular}
};
\node[font=\bfseries] at (4.25,13.2) {Fact Table};

\end{tikzpicture}
}
\caption{A sample chat from our dataset which uses background resources. The chosen spans used in the conversation are shown in \textcolor{blue} {\uline{blue}}. The letters in the brackets denote the type of resource that was chosen - P, C ,R, F and N indicate \textbf{P}lot, \textbf{C}omments, \textbf{R}eview, \textbf{F}act Table and \textbf{N}one  respectively.}
\end{figure*}
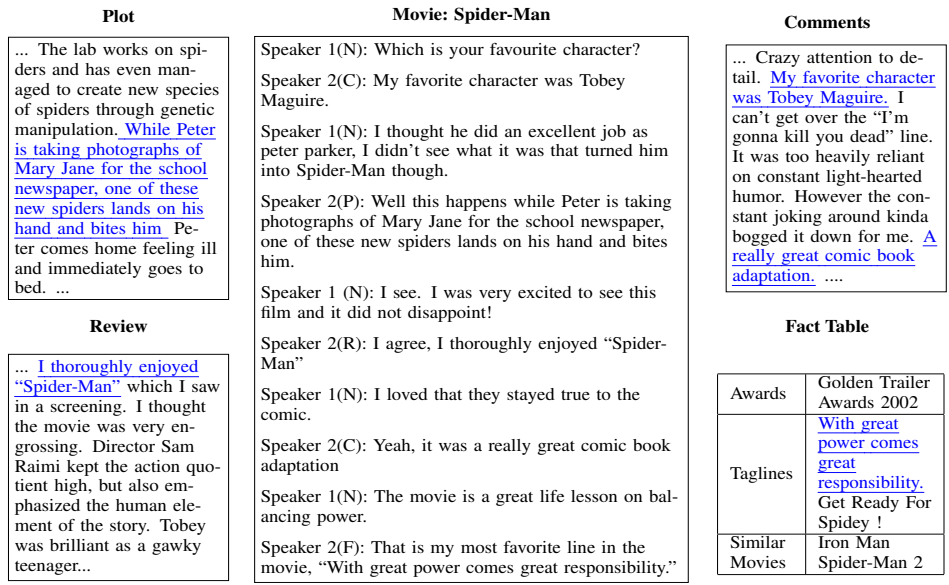

With this motivation, we built a new background aware conversation dataset using crowdsourcing. Specifically, we asked workers to chat about a movie using structured and unstructured resources about the movie such as plots, reviews, comments, fact tables (see Figure \ref{dataset_example}). For every even numbered utterance, we asked the workers to consult the available background knowledge and try to construct a sentence which contains information from this background knowledge and is relevant in the current context of the conversation (akin to how humans recall things from their background knowledge and insert them appropriately in the conversation). For example, in Turn 2, Speaker 2 picked a sentence from the plot which is relevant to the current context of the conversation. Similarly, in Turn 3, Speaker 2 picked a sentence from the movie review. We also asked the workers to suitably modify the content picked from the background knowledge, if needed, so that the conversation remains coherent.  
We collected around 9K such conversations containing a total of 90K utterances pertaining to about 921 movies. These conversations along with the background resources will be made publicly available\footnote{\url{https://github.com/nikitacs16/Holl-E}}. For every utterance, we also provide information about the exact span in the resource from which this utterance was created. Lastly note that unlike existing datasets, our \textit{test set} contains multiple reference responses for each test context thereby facilitating better evaluation of conversation models. We believe that this dataset will allow the community to take a fresh look at conversation modeling and will lead to the development of models which can learn to exploit background knowledge to pick appropriate responses instead of generating responses from scratch. Such a conversation strategy which produces responses from background knowledge would be useful in various domains. For example, a troubleshooting bot could exploit the information available in manuals, reviews and previous bug reports about the software. Similarly, an e-commerce bot could exploit the rich information available in product descriptions, reviews, fact tables, \textit{etc.} about the product. While the proposed dataset is domain specific, it serves as a good benchmark for developing creative background-knowledge-aware models which can then be ported to different domains by building similar datasets for other domains.  

We establish some initial baselines using three different paradigms to demonstrate the various models that can be developed and evaluated using this dataset. For the sake of completeness, the first paradigm is a hierarchical variant of the sequence to sequence architecture which does not exploit any background knowledge. The second paradigm is the copy-and-generate paradigm wherein the model tries to copy text from the given resources whenever appropriate and generate it otherwise. The third paradigm borrows from the span prediction based models which are predominantly being used for Question Answering (QA). These baseline results along with the dataset would hopefully shape future research in the area of background aware conversation models. 

\section{Related Work}

There has been an active interest in building datasets \cite{survey_on_dialog_datasets} for training dialog systems. Some of these datasets contain transcripts of human-bot conversations \cite{DSTC1,DSTC2,DSTC3} while others are created using a fixed set of natural language patterns \cite{weston2017,moviedialog}. The advent of deep learning created interest in the construction of large-scale dialog datasets \cite{ubuntu,ritter2010,sordoninaacl2015} leading to the development of several end-to-end conversation systems \cite{NRM,NCM,personaNCM,HRED} which treat dialog as a sequence generation task.

To make the output of these models more coherent, there is an increasing effort in integrating external background knowledge with these models. This is because human beings rely on background knowledge for conversations as well as other tasks \cite{schallert}. There has been considerable work on incorporating background knowledge in the context of goal-oriented dialog datasets even before the advent of large-scale datasets for deep learning \cite{letsgo,mit_atis}  as well as in recent times \cite{trainable_task_oriented,dst_review,incar} where datasets include small sized knowledge graphs as background knowledge. 
However, the conversations in these datasets are very templated and nowhere close to open conversations in specific domains such as the ones contained in our dataset. 

Even in the case of open domain conversations, there are some works which have integrated external knowledge sources. Most of the entries in 2017 Amazon Alexa Prize \cite{alexa_prize} relied on background knowledge for meaningful response generation. Milabot \cite{milabot} and even the winning entry SoundingBoard \cite{soundingboard}  used Reddit pages, Amazon's Evi Service, and large databases like OMDB, Google Knowledge Graph and Wikidata as external knowledge. The submission named Eigen \cite{eigen} used several dialog datasets and corpora belonging to related Natural Language Processing tasks to make their responses more informative. We refer the reader to \cite{alexa_prize} for detailed analysis of these systems. In the space of academic datasets, \cite{lowe2015} report results on the Ubuntu dataset using manpages as external knowledge whereas \cite{KGNCM} use Foursquare tips as external knowledge for social media conversations. However, unlike our work both these works do not create a new dataset where the responses are explicitly linked to a knowledge source. The infusion of external knowledge in both these works is post facto (as opposed to our work where we take a bottom-up approach and explicitly create a dataset which allows exploitation of background knowledge). Additionally, existing large-scale datasets are noisy as they are extracted from online forums which are inherently noisy. In contrast, since we use crowdsourcing, the extent of noise is reduced since there are humans in the loop who were explicitly instructed to use only clean sentences from the external knowledge sources. 

We would also like to mention some existing works such as \cite{hehe2017,dealnodeal,edina} which have used crowdsourcing for creating conversation datasets. In fact, our data collection method  is inspired by the work of \cite{edina} where the authors use self-dialogs to collect conversation data about movies, music and sports. They are referred to as self-dialogs because the same worker plays the role of both parties in the conversation. However, our work differs from \cite{edina} as we provide explicit background knowledge sources to the workers from where they can copy text with the addition of suitable prefixes and suffixes to generate appropriate coherent responses. 
\section{Dataset}
\label{sect:pdf}
In the following sub-sections we describe the various stages involved in collecting our dataset.

\subsection{Curating a list of popular movies} 

We created a list of 921 movies containing (i) top 10 popular movies within the past five years, (ii) top 250 movies as per IMDb rankings, (iii) top 10 movies in popular genres, and (iv) other popular movie lists made available elsewhere on the Internet. These movies belonged to 22 different genres such as sci-fi, action, horror, fantasy, adventure, romance, \textit{etc.} thereby ensuring that our dataset is not limited to a specific genre. We considered those movies for which enough background information such as plots, reviews, comments, facts, \textit{etc.} were available on the Internet irrespective of whether they were box-office successes or not. Please find the respective urls in the Appendix.

\subsection{Collecting background knowledge} For each movie, we collected the following background knowledge:

\textbf{1. Review (R):} For each movie, we asked some in-house workers to fetch the top 2 most popular reviews for this movie from IMDb using the \textit{sort by Total Votes} option.  We also instructed them to avoid choosing reviews which were less than 50 words but this was typically never the case with popular reviews.
\textbf{2. Plot (P):} For each movie,  
we extracted information about the ``Plot'' of the movie from the Wikipedia page of the movie. Wikipedia pages of movies have an explicit section on ``Plot'' making it easy to extract this information using scripts. 
\textbf{3. Comments (C):}  Websites like \textit{Reddit} have a segment called  ``official discussion page about X'' (where X is a movie name) containing small comments about various aspects of movie. We identified such pages and extracted the first comment on every thread on this page. We bundled all these comments into a single text file and refer to it as the resource containing ``Comments''. For a few movies, the official discussion page was not present in which case we used the review titles of all the IMDb reviews of the movie as comments. The difference between Reviews and Comments  is that a Review is an opinion piece given by one person thus typically exhibiting one sentiment throughout while Comments include opinions of several people about the same movie ensuring that positive, negative and factual aspects of the movie are captured as well as some banter.
 
\textbf{4. Meta data or Fact Table (F):} For each movie, we also collected factual details about the movie, \textit{viz.}, box office collection, similar movies (for recommendations), awards and tag-lines from the corresponding IMDb pages and Wikipedia Infoboxes. Such information would be useful for inserting facts in the conversation, for example, \textit{ ``Did you know that the movie won an Oscar?}''. We included only 4 fields in our fact table instead of showing the entire Wikipedia Infobox to reduce the cognitive load on turkers who already had to read the plot, reviews and comments of the movie.

\subsection{Collecting conversation starters} During our initial pilots, we observed that if we asked the workers to converse for at least 8 turns, they used a lot of the initial turns in greetings and general chit-chat before actually chatting about a movie. To avoid this, we collected opening statements using Amazon Mechanical Turk (AMT) where the task for the workers was to answer the following questions  ``\textit{What is your favorite scene from the movie X ?}'', ``\textit{What is your favorite character from the movie X ?}'' and ``\textit{What is your opinion about the movie X?}'' (X is the movie name). We paid the workers 0.04\$ per movie and showed the same movie to 3 different workers, thereby collecting 9 different opening statements for every movie.  By using these statements as conversation starters in our data collection, the workers could now directly start conversing about the movie.

\subsection{Collecting background knowledge aware conversations via crowdsourcing} 
Our aim is to create a conversation dataset wherein every response is explicitly linked to some structured or unstructured background knowledge. Creating such a dataset using dedicated in-house workers would obviously be expensive and time consuming and so we decided to use crowdsourcing. However, unlike other NLP and Vision tasks, where crowdsourcing has been very successful, collecting conversations via crowdsourcing is a bit challenging. The main difficulty arises from the fact that conversation is inherently a task involving two persons but it is hard to get two workers to synchronize and chat on AMT. We did try a few pilot experiments where we setup a server to connect two AMT workers but we found that the probability of two workers simultaneously logging in was very low. Thus, most workers logged in and left in a few seconds because no other worker joined simultaneously. Finally, we took inspiration from the idea of self chats \cite{edina} in which, the same worker plays the role of both Speaker 1 and Speaker 2 to create the chat. 
In the above self chat setup, we showed every worker 3 to 4 resources related to the movie, \textit{viz.}, plot (P), review (R), comments (C) and fact table (F). We also showed them a randomly selected opening statement from the 9 opening statements that we had collected for each movie and requested them to continue the conversation from that point. The workers were asked to add at least 8 utterances to this initial chat. While playing the role of Speaker 1, the worker was not restricted to copy/modify sentences from the background resources but was given the freedom to create (write) original sentences. However, when playing the role of Speaker 2, the worker was strictly instructed to copy/modify sentences from the shown resources such that they were relevant in the current context of the conversation. The reason for not imposing any restrictions on Speaker 1 was to ensure that the chats look more natural and coherent. Further, Speaker 2 was allowed to add words at the beginning or end of the span selected from the resources to make the chats more coherent and natural (for example, see the prefix in utterance 2 of Speaker 2 in Figure \ref{dataset_example}).  
We paid the workers 40 cents for every chat. Please refer to the Appendix for the instruction screen shots.

\subsection{Verification of the collected chats}

Every chat that was collected by the above process was verified by an in-house evaluator to check if the workers adhered to the instructions and produced coherent chats.  Since humans typically tend to paraphrase the background knowledge acquired by reading articles,  one could argue that such conversations may not look very natural because of this restriction to copy/modify content from the provided resources. To verify this, we conducted a separate human evaluation wherein we asked 15 in-house evaluators to read conversations (without the background resources) from our dataset  and rate them on five different parameters. Specifically, they were asked to check if the conversations were 1) \textbf{intelligible:} \textit{i.e.}, an average reader could understand the conversation 2) \textbf{coherent:} \textit{i.e.}, there were no abrupt context switches 3) \textbf{grammatically correct} 4) \textbf{on-topic:} \textit{i.e.}, the chat revolved around the concerned movie with digression limited to related movies/characters/actors and 5) \textbf{natural two-person chats:} \textit{i.e.}, the role-play setup does not make the chat look unnatural. These evaluators were post-graduate students who were fluent in English and had watched at least 100 Hollywood movies. We did not give them any information about the data creation process. 
We used a total of 500 chats for the evaluation and every chat was shown to 3 different evaluators. The evaluators rated the conversations on a scale of 1 (very poor) to 5 (very good). We computed inter-annotator agreement using the mean linearly weighted Cohen's $\kappa$  \cite{cohen_kappa} and mean Krippendorff's $\alpha$ \cite{kripp_alpha}. The average rating for each of the 5 parameters along with the inter annotator agreement are reported in Table \ref{human_evaluation} and are very encouraging.

\subsection{Statistics}
\begin{table}
\centering
\begin{tabular}{llll}
\hline
\textbf{Metric }   &  \textbf{Rating} & \textbf{$\alpha$} & \textbf{$\kappa$}\\ \hline
\textbf{Intelligible}    & 4.47 $\pm$ 0.52 & 0.70 & 0.69\\ 
\textbf{Coherent}        & 4.33 $\pm$ 0.93 & 0.57 &0.71\\ 
\textbf{Grammar}         & 4.41 $\pm$ 0.56 & 0.60 & 0.69\\ 
\textbf{Two-person-chat} & 4.47 $\pm$ 0.46 & 0.64 &0.70\\ 
\textbf{On Topic}        & 4.57 $\pm$ 0.43 & 0.72 &0.70\\ \hline
\end{tabular}
\caption{Average human evaluation scores with standard deviations for conversations (scale 1-5). We also report mean Krippendorff's $\alpha$ and mean Cohen's $\kappa$}
\label{human_evaluation}
\end{table}
In Table \ref{basic_statistics}, we show different statistics about the dataset collected using the above process. These include average number of utterances per chat, average number of words per utterance, and so on followed by the statistics of the different resources which were used as background knowledge. Please note that the \# unique Plots and \# unique Reviews correspond to unique paragraphs while the \# unique Comments is the count of unique sentences. We observed that 41.2\%, 34.6\%, 16.1\% and 8.1\% of Speaker 2 responses came from Reviews, Comments, Plots and Fact Table respectively.
\begin{table}
\centering
\begin{tabular}{ll}
\hline
\#chats                                 & 9071    \\
\#movies                                & 921     \\
\#utterances                            & 90810   \\
Average \# of utterances per chat       & 10.01  \\
Average \# of words per utterance       & 15.29 \\
Average \# of words per chat            & 153.07 \\
Average \# of words in Plot             & 186.10 \\
Average \# of words in Review           & 384.44 \\
Average \# of words in Comments         & 123.81 \\
Average \# of words in Fact Table       & 33.47   \\
\# unique Plots                         & 5157    \\
\# unique Reviews                       & 1817    \\
\# unique Comments                      & 12740 \\\hline 
\end{tabular}
\caption{Statistics of the dataset}
\label{basic_statistics}
\end{table}

\section{Models}
We evaluate three different types of models as described below. Since these are popular existing models, we describe them very briefly below and refer the reader to the original papers for more details. Note that in this work we merge the comments, reviews, plots and facts into one single document and refer to it as background knowledge. In the rest of the paper, when we refer to a \textit{resource}  we mean this single document which is a merger of all the resources unless specified otherwise.
\subsection{Generation based models} We use the standard Hierarchical Recurrent Encoder Decoder model (HRED) \cite{HRED}  instead of its variant \cite{vhred} as the standard model performs only slightly poorly than the variant and is much easier to implement. It decomposes the context of the conversation as two level hierarchy using Recurrent Neural Networks (RNN). The lower RNN encodes individual utterances (sequence of words) which is then fed into the higher level RNN as a sequence of utterances. The decoder RNN then generates the output based on this hierarchical context representation.   

\subsection{Generate-or-Copy models}  
Get To The Point (GTTP) \cite{GTTP} proposed a hybrid pointer generator network for abstractive summarization that learns to copy words from the source document when required and otherwise generates a word like any sequence-to-sequence model. In the summarization task, the input is a \textit{document} and the output is a \textit{summary} whereas in our case the input is a \{\textit{document, context}\} pair and the  output is a \textit{response}. Here, the context includes the previous two utterances and the current utterance. We modified the architecture to suit our task. We use an RNN to compute the representation of the document (like the original model) and introduce another RNN to compute a representation of the context by treating it as a single sequence of words. The decoder which is also an RNN then uses the document representation, context representation and its own internal state representation to compute a (i) probability score which indicates whether the next word should be copied or generated (ii) probability distribution over the vocabulary if the next word needs to be generated and (iii) probability distribution over the input words if the next word needs to be copied. These three probability distributions are then combined to produce the next word in the response.

\subsection{Span prediction models}Bi-directional Attention Flow Model (BiDAF) \cite{BIDAF} model is a QA model which was proposed in the context of the SQuAD dataset \cite{SQUAD}. Given a \textit{document} and a \textit{question}, the model uses a six-layered architecture to predict the span in the document which contains the answer. We can use their model as it is for our task without any modifications by simply treating the \textit{context} as the \textit{question} and the \textit{resource} as the \textit{document}. 

We chose to evaluate on the modified generate-or-copy model instead of other variants such as \cite{KGNCM,lowe2015} as the modified model already contains the extra encoder for background model which is present in these models. Moreover, the modified model uses a hybrid copy-or-generate decoder which is well-suited to our task. 

\section{Experimental Setup}
In this section we describe the train-validation-test splits, the process used for creating training instances, the manner in which the models were trained using our data and the evaluation metrics. %used for evaluations.

\subsection{Creating train/valid/test splits} On average we have 9.14 chats per movie. We divide the collected chats into train, validation, and test splits such that all the chats corresponding to a given movie are in exactly one of the splits. This ensures that a movie seen in the test or validation set is never seen at training time. We create the splits such that the percentage of chats in the train-validation-test set is roughly 80\%-10\%-10\%. 

\subsection{Creating training instances} For each chat in the training data, we construct training instances of the form \{\textit{resource, context, response}\} where the \textit{context} is taken as previous two utterances and current utterance. We consider only the even numbered utterances as training examples as they are generated from the background resources thus emulating a human-bot setup. If a chat has 10 turns, we will have 5 instances. The task then is to train a model which can predict these even numbered responses. At test time the model is shown \{\textit{resource, context}\} and predicts the response. Note that, HRED will ignore the \textit{resource} and only use \{\textit{context, response}\} as input-output pairs. BiDAF and GTTP will use \{\textit{resource, context, response}\} as training data with relevant \textit{span} instead of \textit{response} for BiDAF.

\subsection{Merging resources into a single document} As stated earlier, we simply merge all the background information to create a single document which we collectively refer to as \textit{resource}. For the BiDAF model, we had to restrict the length of the resource to 256 words because we found that even on a K80 GPU with 12GB RAM, this model gives an out of memory error for longer documents. We found this to be a severe limitation of this and other span based models (for example, R-Net \cite{RNET}) .
	We experimented with three methods of creating this resource. The first method \textit{oracle} uses the actual resource (plot or comments or reviews) from which the next response was generated as a resource. If that resource itself has more than 256 words then we truncate it from the beginning and the end such that the span containing the actual response is contained within the retained 256 words. The number of words that are discarded from the start or the end is chosen at random so that the correct spans do not end up in similar positions throughout the dataset. The next two methods \textit{mixed-short} and \textit{mixed-long} are created by merging the individual resources. We retain each resource in the merged document proportional to its length. (\textit{i.e,}if there are 400 words in the plot, 200 words in the review and 100 in the comments, the merged resource will  contain contiguous sentences from these three resources in the ratio of 4:2:1.) Further, we ensure that the merged resource contains the actual response span. In this way, we create \textit{mixed-short} with 256 words and \textit{mixed-long} with 1200 words (the maximum length of the merged resources).  We will henceforth denote \textit{oracle}, \textit{mixed-long} and \textit{mixed-short} using \lq (o) \rq, \lq (ms) \rq and \lq (ml) \rq respectively. We report results for BiDAF(o), BiDAF (ms), GTTP (o) and GTTP (ml). 

\begin{table*}
\centering
\begin{tabular}{l|cc|cc|cc|cc|cc}
\hline
\textbf{Model}       & \multicolumn{2}{c|}{\textbf{F1}}    &  \multicolumn{2}{c|}{\textbf{BLEU}}  &  \multicolumn{2}{c|}{\textbf{Rouge-1}} & \multicolumn{2}{c|}{\textbf{Rouge-2}} &  \multicolumn{2}{c}{\textbf{Rouge-L}} \\\hline
\textbf{HRED }     & -     & -     & 5.23  &     5.38  & 24.55 & 25.38 & 7.61  & 8.35  & 18.87 & 19.67 \\\hline
\textbf{GTTP (o)   }& -     & -     & 13.92 &      16.46 & 30.32 & 31.6  & 17.78  & 21.21 & 25.67 & 27.83 \\
\textbf{GTTP (ms)}   & -     & -     & 11.05 &      15.68 & 29.66 & 31.71  & 17.70  & 19.72  & 25.13 & 27.35 \\
\textbf{GTTP (ml)}  & -     & -     & 7.51  &    8.73  & 23.20 & 21.55 & 9.91  & 10.42 & 17.35 & 18.12 \\\hline
\textbf{BiDAF (o)}  & 39.69 & 47.18 & 28.85 & 34.98 & 39.68 & 46.49 & 33.72 & 40.58 & 35.91  & 42.64 \\
\textbf{BiDAF (ms)} & 45.73  & 51.35 & 32.95 & 39.39 & 45.69 & 50.73  & 40.18 & 45.01 & 43.46 & 46.95\\\hline

\end{tabular}
\caption{Performance of the proposed models on our dataset. The figures on the left in each column indicate scores on single-reference test dataset while the figures on the right denote scores on multi-reference dataset.}
\label{results}
\end{table*}

\subsection{Evaluation metrics} As HRED and GTTP models are generation based models we use BLEU-4, ROUGE-1, ROUGE-2 and ROUGE-L as the evaluation metrics. For BiDAF we use the above metrics by comparing the predicted span with the reference span. For BiDAF, we also report F1 as stated in \cite{SQUAD}.

In addition to the automatic evaluation, we also collected human judgments using 100 test responses generated for every model for every setup (oracle, mixed-short, mixed-long). These evaluators had the same qualifications as the evaluators who earlier helped us evaluate our dataset. They were asked to rate the response on scale of 1 to 5 (with 1 being the least) on the following four metrics: (1) Fluency(Flu), (2) appropriateness/relevance (apt) of the response in the current context language (3) humanness (Hum) of the response, \textit{i.e.,} whether the responses look as if they were generated by a human (4) and specificity (spec) of the response, \textit{i.e.}, whether the model produced movie-specific responses or generic responses such as ``This movie is amazing''. We report these results in Table \ref{human_evaluation_model}.

\begin{table}[]
\centering
 
\begin{tabular}{l|llll}
\hline
\textbf{Model}     & \textbf{Hum}  &\textbf{Apt}  & \textbf{Flu}   & \textbf{Spec}   \\\hline
\textbf{HRED}   & 3.08 & 2.49          & 2.64 & 2.06 \\
\textbf{GTTP (o)}  & 4.10     & 3.73                                                                                                   &  4.03    & 3.33     \\
\textbf{GTTP (ml)} &2.93	&2.97		&3.42	&2.60 \\
\textbf{BiDAF (o)} & 3.78 & 3.71    & 4.05 & 3.76 \\
\textbf{BiDAF(ms)} & 3.41 & 3.38                                                & 3.47 & 3.30 \\\hline
\end{tabular}
\caption{Human evaluation results on the model performances.}
\label{human_evaluation_model}
\end{table}

\subsection{Collecting multiple reference responses}
One common issue with evaluating dialog systems is that existing datasets typically contain only one reference response whereas in practice several responses can be correct in a given context.  
 To solve this to a certain extent, we collected three reference responses for every Speaker 2 utterance in our dataset (note that Speaker 2 is treated as the bot while training/testing our models). We show the previous utterances ending with Speaker 1's response and ask workers to provide three appropriate responses from the given resources. We found that in some cases there was only one appropriate response like factual response and the workers could not provide multiple references . In this way we were able to create a multiple reference test set where 78.04\% of the test instances have multiple responses. In Table \ref{results}, we report two sets of scores based on single-reference test dataset and multi-reference test dataset. While calculating the scores for multi-reference dataset, we take the maximum score over multiple reference responses.

Please refer to the Appendix section for the details of the model, hyperparameters, example of multiple references in our dataset and sample outputs produced by different models.

\section{Results and Discussion}

In this section, we discuss the results of our experiments as summarized in Tables \ref{results} and \ref{human_evaluation_model}.

\textbf{Generation based models v/s Span prediction models:}
We compare the generation based models and span prediction models only based on results in the \textit{oracle} setting. Here, the span based model (BiDAF) outperforms the generation based models (HRED and GTTP). This confirms our belief that the natural language generation (NLG) capabilities of current generation based models are far from being acceptable even in case of generate-or-copy modes. This also emphasizes the importance of this data which allows building models which can exploit well-formed sentences in the background knowledge and reproduce them with minor modifications instead of generating them from scratch. While the results for BiDAF are encouraging, we reiterate that it does not scale to longer documents (we were not able to run it in the \textit{mixed-long} setting). We still need much better models as BiDAF on SQuAD dataset gives an F1 of 81.52 \% which is much higher than the results on our dataset. Further, note that using the predicted span as a response is not natural. This is evident from human likeliness (Hum) score of GTTP (o) being higher than both the BiDAF models. We need models which can suitably alter the span to retain the coherence of the context.

 \textbf{Effect of including background knowledge:} We observe that there isn't much difference between the performance of HRED which does not use any background knowledge when compared to GTTP (ml) which actually uses a lot of background knowledge. However, there is a substantial difference between the performance of HRED and GTTP (o) which uses only the relevant background knowledge. Further, without background knowledge, HRED learns to produce very generic responses (Spec score = 2.06). This shows that the background knowledge is important, but the models should learn to focus on the right background knowledge relevant to the current context. Alternately, we can have a two-stage network which first predicts the right resource (plot, review, comments) from which the span should be selected and then selects the span from this chosen resource.

 \textbf{\textit{Oracle} v/s \textit{mixed-short} resource:} 
We observe that the performance of BiDAF (ms) is actually better than BiDAF (o) even when the resource length for both is 256 words. We would expect a poor performance for BiDAF (ms) as the resource has more noise because of the sentences from irrelevant resources. However, we speculate the model learns to regard irrelevant sentences as noise and learns to focus on sentences corresponding to the correct resource resulting in improved performance (however, this is only a hypothesis and it needs to be verified). 
We realize that this is clearly a poor baseline and we need better span prediction based models which can work with longer documents. At the same time, GTTP (o) and GTTP (ms) have comparable (yet poor) performance. There is no co-attention mechanism in this model which can effectively filter out noisy sentences.

\textbf{Observations from the copy-and-gen model:}
We observed that this model produced sentences where on average of 82.18\% (\textit{oracle}) and 71.95\% (\textit{mixed-long}) of the tokens were copied. One interesting observation was that it easily learns to copy longer contiguous sequences one word at a time. 
However, as is evident from the automatic evaluation metrics, in many cases, the `copied' spans are not relevant to the current context. 
\\\\
\indent \textbf{Evaluating with multiple references}:
When considering multiple references, the performance numbers as reported in Table \ref{results} indeed improve. This shows the importance of having multiple references and the need to develop metrics which account for multiple dissimilar references.

\section{Conclusion}
We introduce a new dataset for building dialog systems which would hopefully allow the community to take a fresh look at this task. Unlike existing datasets which only contain a sequence of utterances, in our dataset each response is explicitly linked to some background knowledge. This mimics how humans converse by recalling information from their background knowledge and use it appropriately in the context of the conversation. Using  this dataset, we evaluated models belonging to three different paradigms, \textit{viz.}, generation based models, generate-or-copy models and span prediction models. Our results suggest that the NLG capabilities of  existing seq-to-seq models are still far from desirable while span based models which completely bypass the process of NLG show some promise but with clear scope for improvement.

Going forward, we would like to build models which are a hybrid of span prediction models and generation models. Specifically, we would like to build models which can learn to copy a large sequence from the input instead of one word at a time. Another important aspect is to build less complex models which can handle longer documents. For example, the BiDAF model has an expensive outer product between two large matrices which makes it infeasible for long documents (because the size of these matrices grows with the length of the document). Alternately, we would like to build two-stage models which first select the correct resource from which the next response is to be generated and then generate or copy the response from the resource. 
\section*{Acknowledgements}
We would like to thank Department of Computer Science and Engineering, and Robert Bosch Center for Data Sciences and Artificial Intelligence, IIT Madras (RBC-DSAI) for providing us with adequate resources. We also thank Gurneet Singh and Sarath Chandar for helping us in the data collection phase two and three respectively. Lastly, we thank all the AMT workers around the world and our in-house evaluators.
\cleardoublepage
\section*{Appendix}
\section*{Model details - GTTP}
Since we modified the existing architecture of Get to the Point \cite{GTTP}, we now provide details of the same.
In the summarization task, the input is a \textit{document} and the output is a \textit{summary} whereas in our case the input is a \{\textit{resource/document, context}\} pair and the  output is a \textit{response}. Note that the context includes the previous two utterances (dialog history) and the current utterance. Since, in both the tasks, the output is a sequence (\textit{summary} v/s \textit{response}) we don't need to change the decoder (\textit{i.e.}, we can use the decoder from the original model as it is). However, we need to change the input fed to the decoder. Similar to the original model, we use an RNN to compute the representation of the document. Let $N$ be the length of the document then the RNN computes representations $h_1^r, h_2^r, ..., h_N^r$ for all the words in the resource (we use the superscript $r$ to indicate resource). The final representation of the resource is then the attention weighted sum of these word representations:

\begin{equation}
\label{attention_on_document}
\begin{split}
e_i^t &= v^T tanh(W_r h_i^r + U s_t + b_r)\\
a^t &= softmax(e^t)\\
r_t &= \sum_i a_i^t h_i^r
\end{split}
\end{equation}

where $s_t$ is the state of the decoder at the current time step. In addition, in our case, we also have the context of the conversation apart from the document (resource). Once again, we use an RNN to compute a representation of this context. Specifically, we consider the previous $k$ utterances as a single sequence of words and feed these to an RNN. Let $M$ be the total length of the context (\textit{i.e.}, all the $k$ utterances taken together) then the RNN computes representations $h_1^c, h_2^c, ..., h_M^c$ for all the words in the context (we use the superscript $c$ to indicate context). The final representation of the context is then the attention weighted sum of these word representations:

\begin{equation}
\begin{split}
f_i^t &= v^T tanh(W_c h_i^c + V s_t + b_c)\\
m^t &= softmax(f^t)\\
c_t &= \sum_i m_i^t h_i^c
\end{split}
\end{equation}

where $s_t$ is the state of the decoder at the current time step. %Similarly, we also use the document representation to pay attention to important words in the context and compute a modified representation of the context in light of the document 
%
%\begin{equation}
%\textit{This is not document aware query representation. Just an extra encoder in the architecture to suit the query-document-answer format}
%\end{equation}

The decoder then uses $r_t$ (document representation), $c_t$ (context representation) and $s_t$ (decoder's internal state) to compute a probability distribution over the vocabulary $P_{vocab}$. In addition the model also computes $p_{gen}$ which indicates that there is a probability $p_{gen}$ that the next word will be \textit{generated} and a probability $(1 - p_{gen})$ that the next word will be \textit{copied}. We use the following modified equation to compute $p_{gen}$
\begin{equation}
p_{gen} = \sigma(w_{r}^T r_t + w_{c}^T c_t + w_s^T s_t + w_x^T x_t + b_{g})
\end{equation}

where $x_t$ is the previous word predicted by the decoder and fed as input to the decoder at the current time step. Similarly, $s_t$ is the current state of the decoder computed using this input $x_t$. The final probability of a word $w$ is then computed using a combination of two distributions, \textit{viz.}, ($P_{vocab}$) as described above and the attention weights assigned to the document words as shown below

%We do not alter the equation to select words from the extended vocabulary distribution
\begin{equation}
P(w) = p_{gen} P_{vocab}(w) + (1 - p_{gen}) \sum_{i:w_i=w} a_i^t
\end{equation}
where $a_i^t$ are the attention weights assigned to every word in the document as computed in Equation \ref{attention_on_document}. Thus, effectively, the model could learn to copy a word $i$ if $p_{gen}$ is low and $a_i^t$ is high. 
 
\section*{Example from the multiple reference test set}
As seen from the Table \ref{multi_test}, the given chat on ``Secret Life of Pets'' can have multiple responses for Speaker 2. Notice how Reference 1 talks against low critique scores thus emphasizing that he was totally impressed by the movie while Reference 4 has neutral opinion about the same. At the same time, Reference 3 talks about movie specific details like his favorite character while Reference 4 gives a personal opinion. All these four responses are valid given the current context.
\begin{table*}[]
\centering
\begin{tabular}{|l|l|}
\hline
Movie  Name & The Secret Life of Pets                                                                                                                                                                    \\ \hline
Chat        & \begin{tabular}[c]{@{}l@{}}Speaker1 : What do you think about the movie? \\ Speaker2 : I think it was comical and entertaining. \\ Speaker1 : It delivered what was promised.\end{tabular} \\ \hline
Reference 1 & I agree! I'm surprised this film got such a low overall score by users.                            \\ \hline
Reference 2 & My favorite character was Gidget!  She was so much fun and so loyal to her friends!                                                                                                        \\ \hline
Reference 3 & \begin{tabular}[c]{@{}l@{}}Yes! As a Great Dane owner, I often wonder what my dogs are thinking. \\ It was fun to see this take on it.\end{tabular}                                        \\ \hline
Reference 4 & It was full of cliches with a predictable story, but with some really funny moments.                                                                                                       \\ \hline
\end{tabular}
\caption{Multiple references for the given chat.}
\label{multi_test}

\end{table*}
\section*{Hyper-parameters}
We describe the hyperparameters that we used for each model in this sub section. Following the original paper, we trained the HRED model using Adam \cite{adam} optimizer with an initial learning rate of 0.001 on a minibatch of size 16. We used a dropout \cite{dropout} with a rate of 0.25. For word embeddings we use pre-trained GloVe \cite{glove} embeddings of size 300. For all the encoders and decoders in the model we used Gated Recurrent Unit (GRU) with 300 as the size of the  hidden state. We restricted our vocabulary size to 20,000 most frequent words.

We followed the hyperparameters mentioned in the original paper and trained GTTP using Adagrad \cite{adagrad} optimizer with an initial learning rate of 0.15 and an initial accumulator value of 0.1 on a minibatch of size 16. For the encoders and decoders we used LSTMs with 256 as the size of the  hidden state. To avoid vanishing and exploding gradient problem we use gradient clipping with a maximum gradient norm of 2. We used early stopping based on the validation loss. %We restricted our vocabulary size to 30,000.\\

Again following the original paper, we trained BiDAF using AdaDelta \cite{adadelta} optimizer with an initial learning rate of 0.5 on a minibatch of size 32. For all encoders, we use LSTMs with 256 as the size of the  hidden state. We used a dropout \cite{dropout} rate of 0.2 across all LSTM layers, and for the linear transformation before the softmax for the answers. For word embeddings we use pre-trained GloVe embeddings of size 100. For both GTTP and BiDAF, we had to restrict context length to 65 tokens for fair comparison. Note that GTTP can scale beyond 65 tokens but BiDAF cannot.

\section*{Sample responses produced by the models}
As seen from Table \ref{example-chat}, HRED isn't able to produce responses that correspond to the given movie or the given context as it lacks any notion of background knowledge associated with it. We will not consider HRED for the following discussion. In Example 1, we can clearly see that only GTTP (oracle) matches with the ground truth. The remaining three models produce varied outputs which are still relevant to the context. In Example 2, we observe that prediction based models produce appropriate recommendation because of better context-document mapping mechanisms. Both the GTTP models produce responses which are copied but irrelevant to the context. At the same time, just producing spans without any structure isn't natural. This explains the need for hybrid models. Example 3 asks for the backstory of a character which requires complex reasoning. The model has to first understand the plot of the movie to locate the sentences which talk about that character's past. As seen from the responses of the given models, all of them except GTTP (ml) pick sentences which are relevant to the character but do not answer the required question. As discussed earlier, these models rely on word-overlap and thus possess limited natural language understanding. Thus, we need models which are capable of going beyond word overlap and producing responses even in such complex scenarios.
\begin{table*}
\centering
\begin{tabular}{|l|p{4cm}|p{4cm}|p{4cm}|}
\hline
           Movie   & Example 1  (Creed)                                                                                                                                                                                                                                                                                                                                                                                                                                                                                 & Example 2 (Ted)                                                                                                                                                                                                                                                                                                                                                                                                                                                                                                                                                                                                                                                                                                                                                                                                                                                                                                                                                                                                                                                                                                           & Example 3 (Jaws)                                                                                                                                                                                                                          \\ \hline
Context & 
S1 : Which was the best scene in the movie according to you? \newline S2 : I liked the one in which Rocky trains and runs.\newline S1 : It was great when the match commences, and although Donnie takes a few hits, he defeats Leo. \newline S2 : Creed goes the distance as an exceptional crowd pleaser. \newline S1  : I agree with grit, style and substance. There's no other way to say it; Creed is a knockout from start to finish. & S2 : The scene where they're yelling at each other out the window was gold \newline S1 : Thats a similar scene from family guy. I couldnt stop laughing.  \newline S2 : I thought every scene was gold. I also couldn't believe Liam Neeson was in the movie. \newline S1 : I will love to pick something similar this evening. Do you have a recommendation? & S1 : What do you think about the characters in this movie? \newline S2 : My favorite character was Brody, he was such a fun character.\newline S1 : I didn't get Quint's backstory. \\ \hline
Ground Truth  & It wholeheartedly captures what was so brilliant about the first film: the characters.                                                                                                                                                                                                                                                                                                                                                                                                      & Yes. You can try 22 Jump Street.                                                                                                                                                                                                                                                                                                                                                                                                                                                                                                                                                                                                                                                                                                                                                                                                                                                                                                                                                                                                                                                                                     & Quint, aboard that ship in World War II when it was sunk by torpedoes, had witnessed the deaths of roughly 800 men that day, many of whom were eaten by sharks as they struggled helplessly in the water.                           \\ \hline
HRED          & yeah , but as the ark is launched , and you are right taxi driver , runs with the beauty , \textless{}UNK\textgreater \textless{}UNK\textgreater and \textless{}UNK\textgreater \textless{}UNK\textgreater .                                                                                                                                                                                                                                                                                & the plot is a classic in the film , it 's not that it 's an original and cinematic experience .                                                                                                                                                                                                                                                                                                                                                                                                                                                                                                                                                                                                                                                                                                                                                                                                                                                                                                                                                                                                                      & i 'm not ashamed to admit it , mary poppens is bangin ! ! !                                                                                                                                                                         \\ \hline
BiDAF(o) & I can't say enough great things about writer / director Ryan Coogler.                                                                                                                                                                                                                                                                                                                                                                                                                       & The Hangover                                                                                                                                                                                                                                                                                                                                                                                                                                                                                                                                                                                                                                                                                                                                                                                                                                                                                                                                                                                                                                                                                                         & Hooper notices the shark beginning to circle the boat, and Quint rushes out for a look.                                                                                                                                             \\ \hline
BiDAF(ms)  & I loved how they introduced each opponent with his record and accomplishments.                                                                                                                                                                                                                                                                                                                                                                                                              & 21 Jump Street                                                                                                                                                                                                                                                                                                                                                                                                                                                                                                                                                                                                                                                                                                                                                                                                                                                                                                                                                                                                                                                                                                       & The mood suddenly darkens as Quint admits that the former tattoo represented the US Navy cruiser                                                                                                                                    \\ \hline
GTTP(o)  & it wholeheartedly captures what was so brilliant about the first film : the characters .                                                                                                                                                                                                                                                                                                                                                                                                    & \$ 218,628,680 i think .                                                                                                                                                                                                                                                                                                                                                                                                                                                                                                                                                                                                                                                                                                                                                                                                                                                                                                                                                                                                                                                                                             & brody is stunned and alerts quint .                                                                                                                                                                                                 \\ \hline
GTTP(ml)    & i can't say enough great things about writer / director ryan coogler .                                                                                                                                                                                                                                                                                                                                                                                                                      & this movie had way too much product placement .                                                                                                                                                                                                                                                                                                                                                                                                                                                                                                                                                                                                                                                                                                                                                                                                                                                                                                                                                                                                                                                                      & jaws' is the original summer blockbuster                                                                                                                                                                                            \\ \hline
\end{tabular}
\caption{Examples produced by various models. S1 denotes Speaker1 while S2 denotes Speaker2. \lq o\rq,\lq ml\rq ~and \lq ms\rq ~represent \textit{oracle}, \textit{mixed-long} and \textit{mixed-short} versions of the dataset respectively.}
\label{example-chat}

\end{table*}
\if 0
\section*{Collection of popular movies list}
Following are the urls used to curate the popular movies list:\\
IMDb top 250 :\\ \url{https://www.imdb.com/chart/top}
Popular movies by genre: \\
\url{https://www.imdb.com/feature/genre/}\\
Other popular movies lists: \\
\url{http://www.filmsite.org/guinness.html}\\
\url{https://www.thetoptens.com/best-movie-genres/}\\
\url{http://1001films.wikia.com/wiki/The_List}\\
The imdb ids of the movies will be made available with every example.
\fi
\section*{Data Collection Interfaces used on Amazon Mechanical Turk}
\begin{figure*}
\centering
\resizebox{\textwidth}{8cm}{
\includegraphics{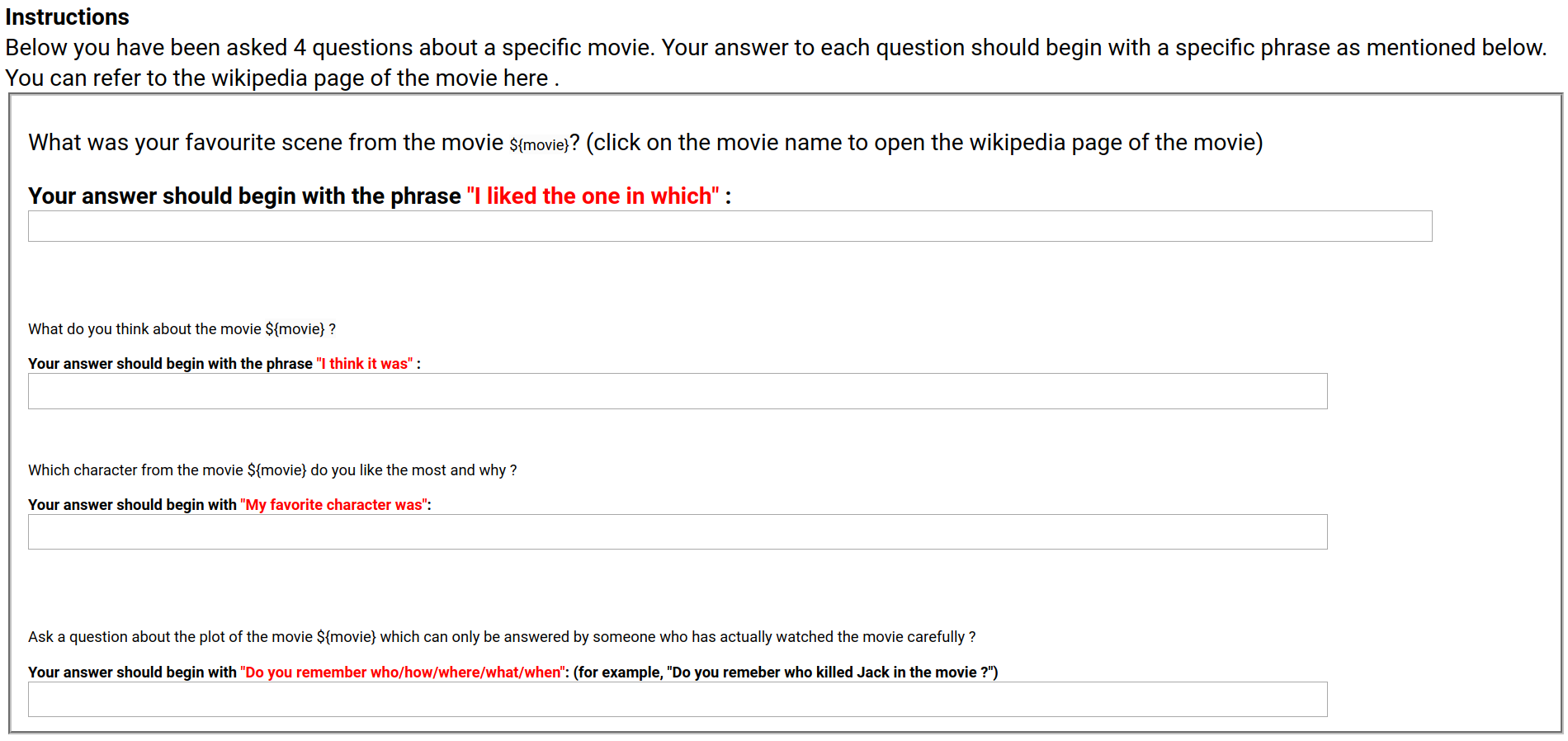}

}
\caption{Instruction screen for opening statement collection from the stage of the dataset collection procedure}
\label{phase_3}
\end{figure*}

As explained in the paper, we resorted to AMT for three types of data collection \textit{viz.} collection of opening statements as discussed in stage 3 of the dataset collection (Figure \ref{phase_3}), the actual chat data collection explained in stage 4 of the procedure and additional responses collected for the test dataset. We will now show the instruction screens. The html files for the same will be released along with the code.
\begin{figure*}
\centering
\resizebox{7.5cm}{11.2cm}{
\includegraphics{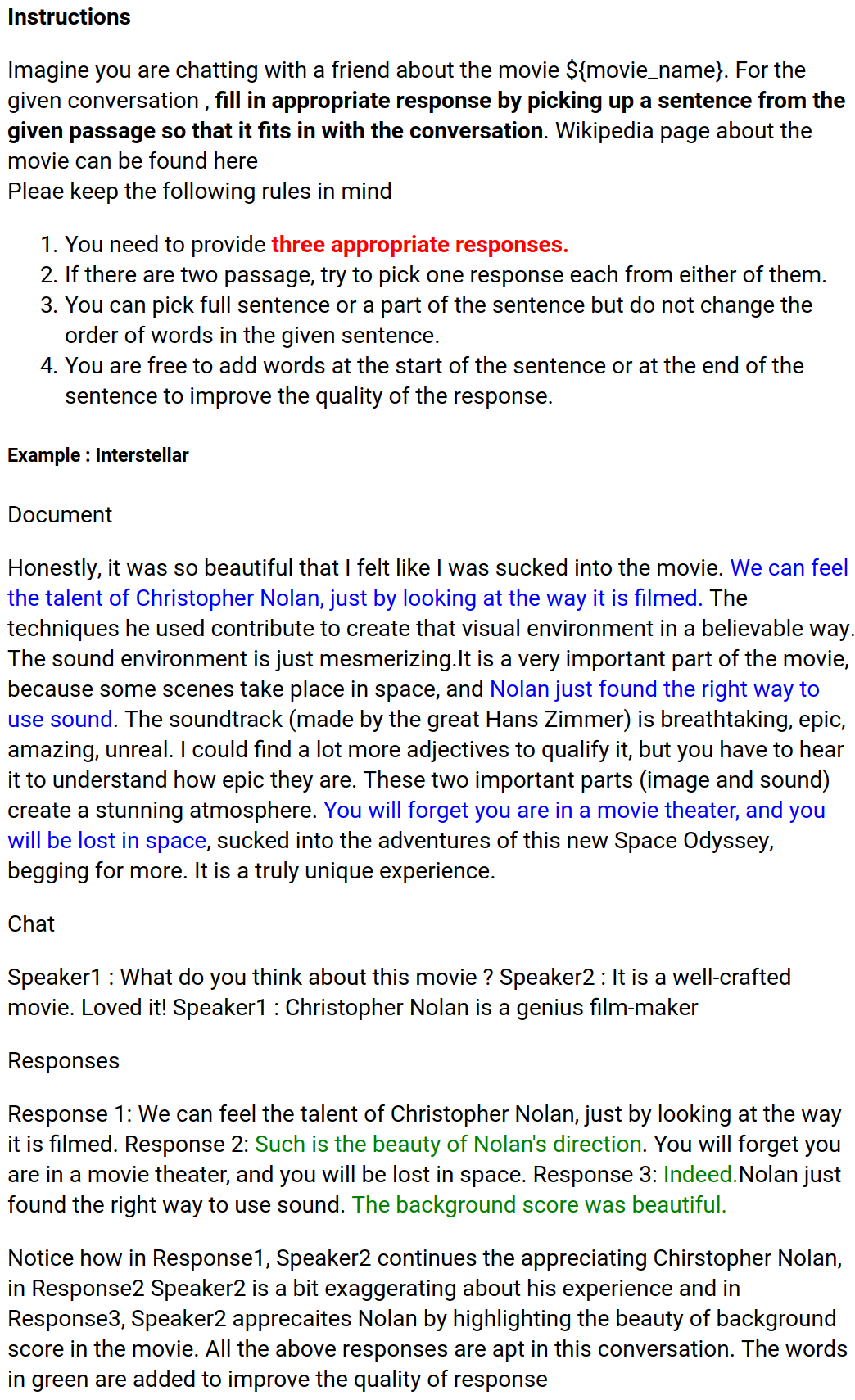}
}
\caption{Instruction screen for collection of multiple-responses for the same chat for the test dataset.}
\label{multi-response.png}
\end{figure*}
\begin{figure*}
\centering
\resizebox{\textwidth}{8cm}{
\includegraphics{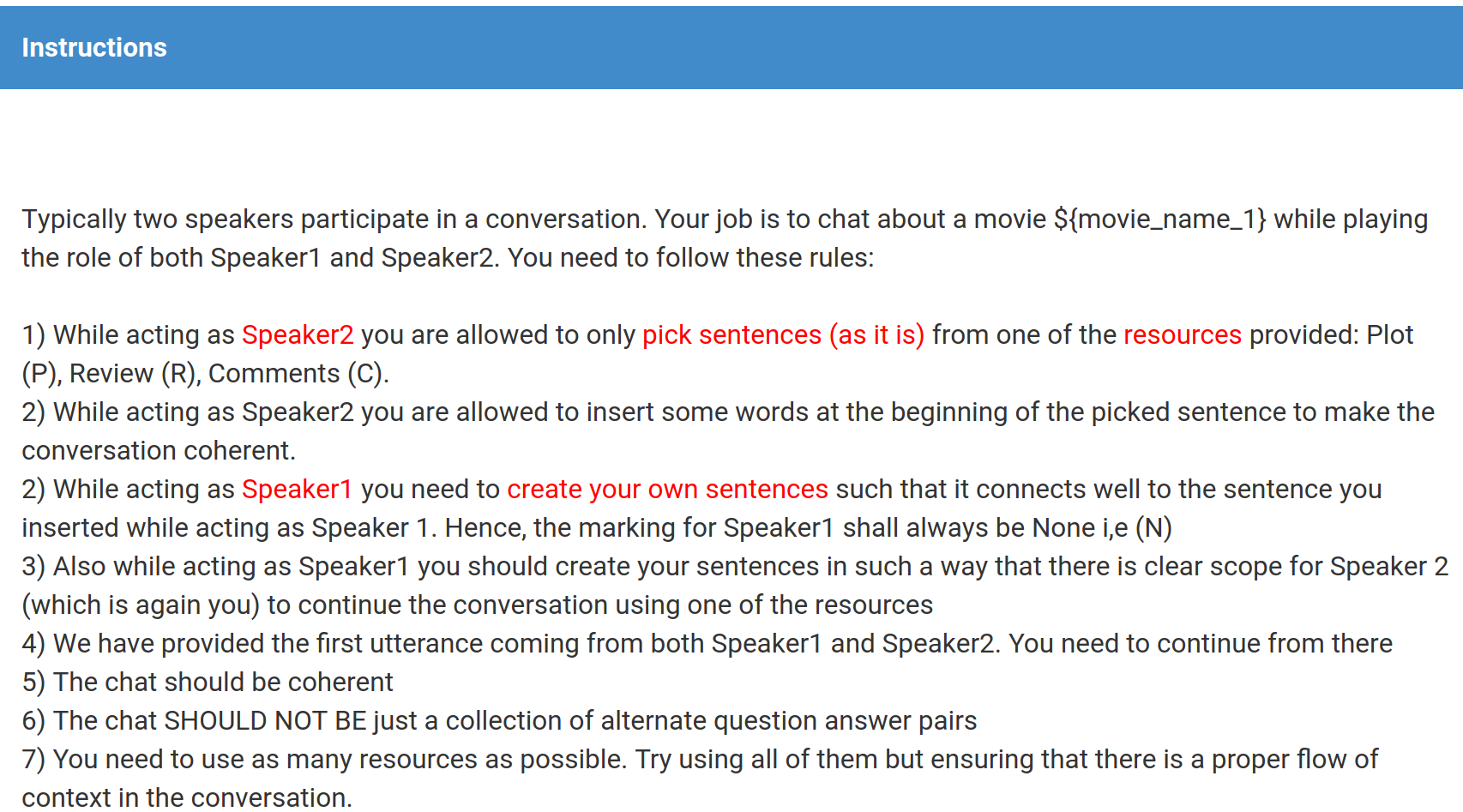}
\label{}
}
\caption{Instruction screen for chat data collection from Phase 4 of the dataset collection procedure}
\end{figure*}
\begin{figure*}
\centering
\resizebox{9.5cm}{13cm}{
\includegraphics{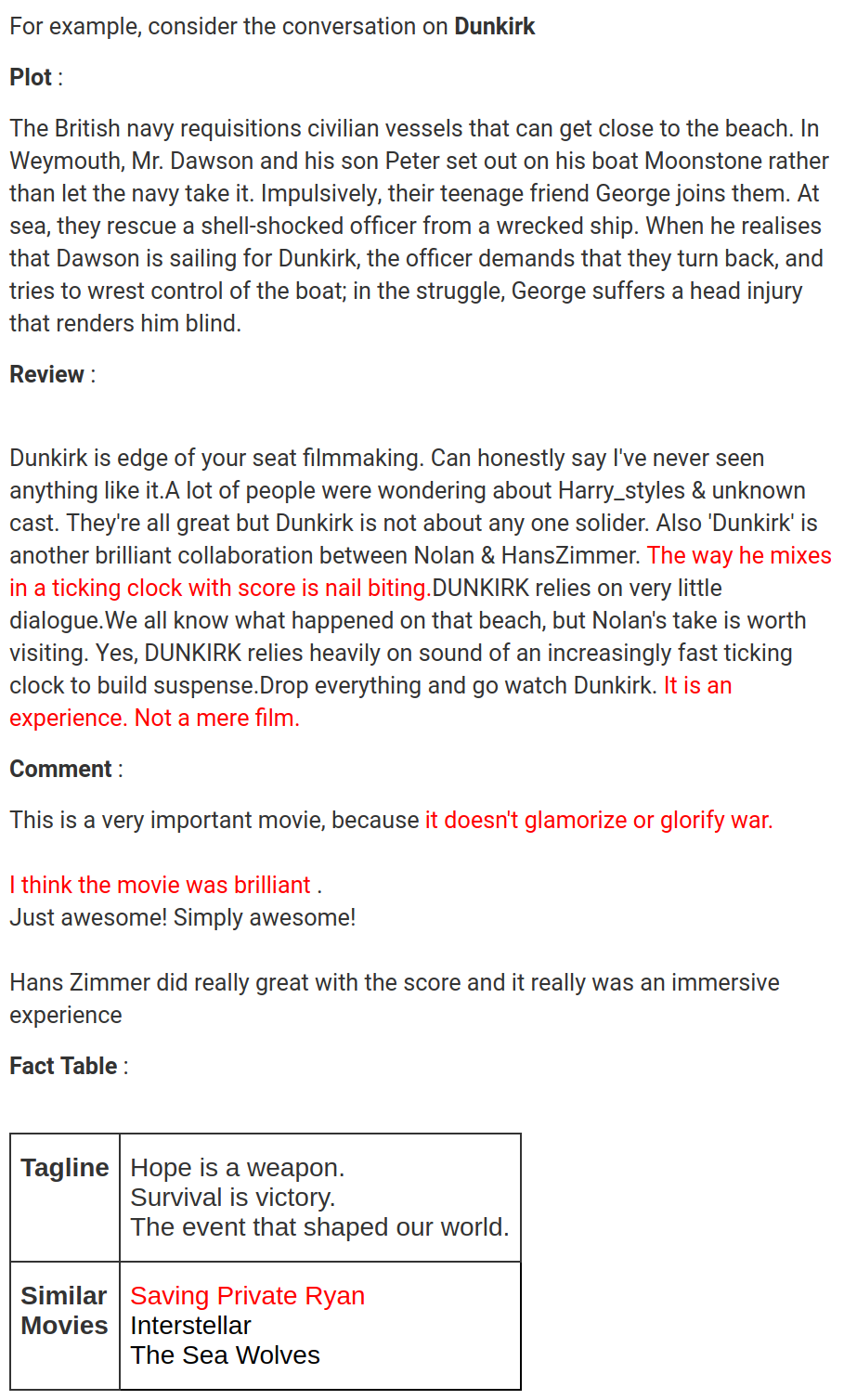}
}

\caption{Background resources for the example chat shown to the workers on AMT}
\end{figure*}
\begin{figure*}
\centering

\includegraphics[scale=0.3]{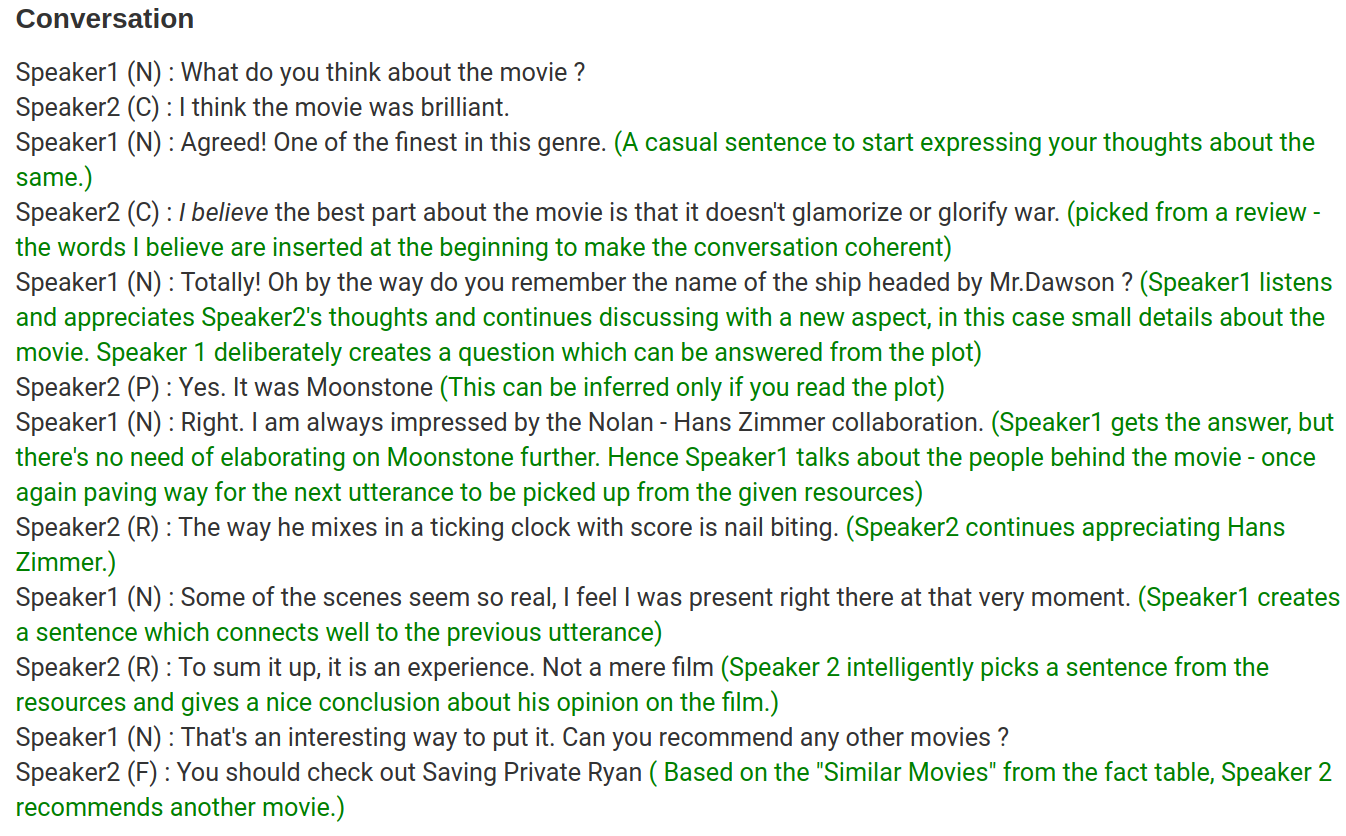}

\caption{Example chat shown to the workers on AMT}
\end{figure*}

%\cleardoublepage
\newpage
\bibliography{emnlp}
\bibliographystyle{acl_natbib_nourl}
\end{document}